\documentclass[
hf,
]{ceurart}

\usepackage{hyperref}       
\usepackage{url}            
\usepackage{booktabs}       
\usepackage{amsfonts}       
\usepackage{nicefrac}       
\usepackage{microtype}      
\usepackage{xcolor}         

\usepackage{graphicx}
\usepackage{amsmath}
\usepackage{amssymb}
\usepackage{array}
\usepackage{multirow}
\usepackage{color}
\usepackage{colortbl}
\usepackage{framed}
\usepackage{bm}
\usepackage{xspace}
\usepackage{enumitem}
\usepackage{xparse}
\usepackage{algorithm}
\usepackage{listings}
\usepackage{url}
\usepackage{mathtools}
\usepackage{pifont}
\usepackage{mathtools}
\usepackage{lipsum}
\usepackage{adjustbox}
\usepackage{bbm}
\usepackage{wrapfig}
\usepackage{algorithm}
\usepackage{algpseudocode}

\newcommand{\cmark}{\ding{51}}
\newcommand{\xmark}{\ding{55}}

\definecolor{darkred}{RGB}{212, 0, 0}
\definecolor{darkgreen}{RGB}{0, 128, 0}
\definecolor{Gray}{gray}{0.2}
\definecolor{lightgray}{gray}{0.92}
\definecolor{blond}{rgb}{0.98, 0.94, 0.75}
\definecolor{TitleColor}{gray}{0.95}
\definecolor{LightCyan}{rgb}{0.88,0.95,1}
\definecolor{OurColor}{rgb}{0.855, 0.937, 0.957}

\definecolor{blond}{rgb}{0.98, 0.94, 0.75}

\def \eg {\emph{e.g.}}

\newcommand{\tit}[1]{\smallskip\noindent\textbf{#1.}}

\usepackage{listings}
\begin{document}
\copyrightyear{2026}
\copyrightclause{Copyright for this paper by its authors.
  Use permitted under Creative Commons License Attribution 4.0
  International (CC BY 4.0).}

\conference{ }

\title{Diffusion Language Models: An Experimental Analysis}


\author[1]{Thomas Bertolani}[%
orcid=0009-0005-3638-7862,
email=282884@studenti.unimore.it,
]
\cormark[1]
\address[1]{University of Modena and Reggio Emilia, Italy}
\address[2]{University of Pisa, Italy}

\author[1,2]{Davide Bucciarelli}[%
orcid=0009-0002-9652-8311,
email=davide.bucciarelli@unimore.it,
]

\author[1]{Leonardo Zini}[%
orcid=0009-0003-9439-9867,
email=leonardo.zini@unimore.it,
]
\author[1]{Marcella Cornia}[%
orcid=0000-0001-9640-9385,
email=marcella.cornia@unimore.it,
]

\author[1]{Lorenzo Baraldi}[%
orcid=0000-0001-5125-4957,
email=lorenzo.baraldi@unimore.it,
]

\cortext[1]{Corresponding author.}

\begin{abstract}
Large Language Models (LLMs) have revolutionized language modeling through autoregressive generation, enabling strong performance across a wide range of tasks. Recently, Diffusion Language Models (DLMs) have emerged as an alternative paradigm that generates text through iterative denoising rather than next-token prediction, allowing parallel refinement of entire sequences. While numerous diffusion-based architectures have been proposed, differences in evaluation protocols, datasets, inference budgets, and generation hyperparameters make it difficult to compare their capabilities and understand the trade-offs they offer. In this work, we present a systematic experimental analysis of modern DLMs. Specifically, we evaluate eight state-of-the-art DLMs across eight benchmarks spanning reasoning, coding, translation, knowledge, and structured problem solving, while explicitly considering both generation quality and computational efficiency. Beyond downstream evaluation, we analyze the impact of key inference-time factors, including denoising steps, context length, block size, and parallel unmasking strategies, and complement large-scale experiments with controlled comparisons of smaller models trained under identical conditions. Our analysis highlights the strengths and limitations of diffusion-based language modeling across different tasks, architectures, and inference budgets. We show that the behavior of DLMs is strongly influenced by generation-time design choices, leading to distinct trade-offs between performance and computational efficiency. Overall, our study provides practical insights into the capabilities and deployment characteristics of contemporary DLMs.
\end{abstract}

\begin{keywords}
  Diffusion Language Models \sep
  Experimental Analysis \sep
  Large Language Models \sep
  Diffusion Models \sep
  Non-Autoregressive Models
\end{keywords}

\maketitle

\section{Introduction}
Large Language Models (LLMs) are predominantly based on autoregressive generation~\cite{grattafiori2024llama, yang2025qwen3, radford2019language, team2024gemma}, where text is produced sequentially one token at a time. While highly effective, this paradigm imposes a strict left-to-right dependency during inference, limiting opportunities for parallel generation and global refinement of generated content. These limitations have motivated the exploration of alternative generation paradigms that can leverage parallelism and iterative refinement during decoding~\cite{leviathan2023fast, chen2023accelerating}.

Diffusion Language Models (DLMs) have recently emerged as a promising alternative~\cite{arriola2025block,arriola2026encoder,sahoo2024simple, sahoo2025diffusion,ye2025dream,zhu2025llada, nie2026large}. Instead of generating text token-by-token, DLMs formulate generation as an iterative denoising process that progressively transforms a corrupted sequence into coherent text. Inspired by the success of diffusion models in domains such as image, video, and audio generation~\cite{dhariwal2021diffusion,rombach2022high,esser2024scaling,kong2020diffwave}, a growing body of work has adapted diffusion techniques to language, first through continuous latent representations and more recently through discrete token-level formulations. These approaches offer several attractive properties, including bidirectional context modeling, parallel token refinement, and the ability to naturally support tasks such as infilling, editing, and globally constrained reasoning.

The rapid development of diffusion language modeling has produced a diverse ecosystem of architectures. Recent proposals span fully discrete diffusion models, hybrid encoder-decoder formulations, and block-diffusion approaches that combine autoregressive conditioning with local diffusion-based refinement. While these models have demonstrated increasingly competitive performance, their evaluation remains highly fragmented. Individual works are often assessed on different benchmarks, using distinct generation budgets, sampling schedules, context lengths, and inference configurations. As a result, comparing results across papers is difficult, and it remains unclear whether reported gains stem from architectural improvements or from differences in evaluation protocols. Furthermore, one of the defining characteristics of diffusion-based language generation is the explicit trade-off between quality and computational cost. Unlike autoregressive models, whose generation procedure is largely fixed, diffusion models expose several inference-time parameters (\eg, the number of denoising steps, sequence length, block size, and unmasking schedules) that directly influence both performance and efficiency. Despite the importance of these factors, their impact has not yet been systematically characterized across modern diffusion architectures.

To address this gap, we present a comprehensive experimental analysis of state-of-the-art DLMs. We evaluate representative pure diffusion and block-diffusion architectures under a unified protocol and compare them against strong autoregressive baselines. Our study examines performance across general knowledge~\cite{gema2025we,hendrycks2020measuring}, reasoning~\cite{cobbe2021training,ye2025beyond,zellers2019hellaswag}, coding~\cite{chen2021evaluating,austin2021program}, and machine translation~\cite{bojar2016findings} benchmarks, while explicitly analyzing the trade-offs between quality and computational efficiency. Beyond downstream evaluation, we conduct controlled scaling experiments to quantify the effects of denoising schedules, sequence lengths, block sizes, and parallel unmasking strategies on model behavior. We also investigate the practical computational requirements of these architectures, providing insights into their memory consumption and inference costs under different generation settings.

Our contributions can be summarized as follows: \textit{(i)} We provide a unified evaluation of modern DLMs across a diverse set of downstream benchmarks, enabling direct comparison between competing architectural paradigms; \textit{(ii)} We systematically analyze the quality--efficiency trade-offs induced by key diffusion hyperparameters, including denoising steps, context length, block size, and unmasking ratios; \textit{(iii)} We complement large-scale downstream evaluations with controlled small-scale experiments, allowing us to study architectural properties through perplexity and scaling analyses under identical training conditions; \textit{(iv)} We provide a comparative analysis of the computational requirements of DLMs, reporting peak memory usage and floating-point operations for both single forward passes and complete generation, highlighting the practical deployment trade-offs between pure diffusion and block-diffusion architectures.

\section{Background}

\subsection{Autoregressive Language Modeling}

Autoregressive language models define a probability distribution over token sequences by factorizing the joint distribution into a product of conditional distributions:
\begin{equation}
p(x_1, \ldots, x_T) = \prod_{t=1}^{T} p(x_t \mid x_{<t}).
\end{equation}

This formulation enables straightforward maximum likelihood training and has scaled effectively with Transformer architectures, leading to strong empirical performance across a wide range of natural language processing tasks~\cite{austin2021program,caffagni2024revolution,zini2026vhector,bucciarelli2024personalizing}. In particular, large-scale autoregressive language models exhibit emergent zero-shot and in-context learning capabilities, making them the dominant paradigm for language modeling.

Despite these strengths, autoregressive generation is inherently sequential: tokens must be generated one at a time conditioned on previous context. This limits parallelism during inference and results in latency that grows linearly with sequence length. Additionally, the left-to-right generation process requires models to commit to decisions sequentially, making it difficult to revisit earlier predictions once produced. While these limitations have not prevented autoregressive models from achieving state-of-the-art performance, they motivate the exploration of alternative generation paradigms that may offer improved efficiency or different quality--compute trade-offs.

\subsection{Diffusion as a Generative Paradigm}
Diffusion models provide an alternative approach to generative modeling based on iterative denoising. A data sample is gradually corrupted through a forward noising process, and a neural network is trained to reverse this process step by step~\cite{esser2024scaling,bucciarelli2026tiny}. Generation is performed by starting from noise and iteratively refining the sample into a coherent output.
A key conceptual advantage of diffusion models is their ability to refine an entire representation in parallel at each denoising step. Rather than progressively extending a partial output, the model repeatedly updates all components of the sample simultaneously, whether they correspond to image latents or textual tokens. This global refinement process allows information to propagate across the entire representation throughout generation, potentially enabling more coherent outputs and greater parallelism than autoregressive decoding.
However, diffusion models introduce a different computational bottleneck: generation typically requires multiple denoising iterations. As a result, inference efficiency depends not only on the computational cost of a single forward pass but also on the total number of refinement steps, creating an explicit trade-off between generation quality and computational efficiency.

\subsection{Continuous Diffusion for Language}
Early attempts to apply diffusion models to language modeling extended continuous diffusion techniques from vision to text by operating in a continuous embedding space. In these approaches, discrete tokens are mapped into continuous vector representations, and Gaussian noise is applied in this latent space. A neural network is then trained to denoise corrupted embeddings.

This design enables the reuse of standard diffusion mechanisms developed for continuous data domains and allows for parallel updates across the sequence. However, several fundamental challenges arise due to the mismatch between continuous noise processes and discrete linguistic structure.

First, language is inherently discrete, and perturbations in embedding space do not correspond to well-defined symbolic transformations. As a result, the denoising trajectory may traverse regions of the embedding space that do not map cleanly to valid or meaningful tokens. Second, the final step of projecting continuous representations back into discrete tokens introduces additional quantization error and can degrade generation quality. Finally, small perturbations in embedding space can induce disproportionately large semantic changes, making the denoising process difficult to stabilize.

These issues suggest that continuous diffusion may be misaligned with the discrete nature of language, motivating approaches that define diffusion processes directly over token sequences.

\subsection{Discrete Diffusion Language Models}
Discrete diffusion models address the limitations of continuous approaches by operating directly in token space. Instead of adding Gaussian noise to embeddings, tokens are progressively corrupted through discrete stochastic processes, such as masking, token replacement, or categorical noise injection. A neural model is trained to reconstruct the original sequence from these corrupted versions.

Two common corruption strategies are \emph{uniform} and \emph{absorbing} diffusion. In uniform diffusion, tokens are replaced with randomly sampled vocabulary items, gradually transforming the sequence toward a uniform distribution over tokens. In absorbing diffusion, tokens are progressively replaced by a special absorbing state (\eg, a \texttt{[MASK]} token), allowing the model to learn reconstruction from partially masked inputs. These choices influence both the learning dynamics and the generation behavior of the resulting models. In practice, absorbing diffusion has become the predominant approach in recent DLMs, as empirical studies have generally found it to provide stronger performance and more stable training dynamics than uniform corruption schemes~\cite{austin2021structured}.

This formulation aligns naturally with language structure and removes the need for continuous-to-discrete output projection. Moreover, discrete diffusion enables parallel prediction of multiple tokens during each denoising step, offering a potential efficiency advantage over autoregressive generation.

Like other diffusion-based approaches, discrete diffusion models rely on an iterative denoising process at inference time, making generation quality and computational efficiency closely linked to the number of denoising steps performed. While reducing the number of steps can accelerate generation, it may also affect output quality, motivating research into architectures and sampling strategies that better balance these objectives. Despite this trade-off, recent advances have shown that discrete diffusion models can achieve competitive performance across a variety of language generation tasks.

\subsection{Block and Hybrid Diffusion Approaches}
To address the inference costs associated with iterative diffusion-based generation, recent work has explored hybrid architectures that combine autoregressive and diffusion-based modeling. 

Rather than generating a sequence token-by-token as in conventional autoregressive language models, these approaches partition the output into blocks of tokens. Generation then proceeds autoregressively at the block level: each block is conditioned on previously generated blocks, preserving a causal structure that facilitates long-range coherence and efficient context modeling.
Within each block, however, tokens are generated using a diffusion process rather than a left-to-right decoding procedure. The model iteratively denoises the tokens of a block in parallel, allowing local refinement and reducing the sequential dependencies that characterize autoregressive decoding. Consequently, block diffusion can be viewed as a compromise between the two paradigms: autoregressive generation provides global structure across blocks, while diffusion-based generation enables parallel refinement within each block.

By combining these mechanisms, block diffusion aims to achieve a more favorable quality--efficiency trade-off than either paradigm alone. Compared to autoregressive models, it reduces sequential decoding steps during inference, while avoiding full-sequence iterative refinement typical of diffusion models.
Nevertheless, its effectiveness depends on the chosen block structure and denoising schedule, and the balance between autoregressive and diffusion components remains an active area of research.

\section{Related Work}

\subsection{Generative paradigms}

\tit{Autoregressive Language Models}
Autoregressive language models form the standard paradigm for neural text generation~\cite{radford2019language, brown2020language, grattafiori2024llama, team2024gemma, guo2025deepseek, yang2025qwen3}, originating from early neural language models and scaling through Transformer-based architectures~\cite{vaswani2017attention}. These models have demonstrated strong scaling behavior and emergent in-context learning abilities at large scale~\cite{kaplan2020scaling}.
In our experiments, we adopt GPT-2~\cite{radford2019language} and Qwen3~\cite{yang2025qwen3} as representative autoregressive baselines against which DLMs are evaluated.
However, their strictly sequential decoding process has motivated a long line of work on improving inference efficiency, including speculative decoding, caching strategies, and parallel decoding approximations~\cite{pope2023efficiently, leviathan2023fast, chen2023accelerating}. Despite these efforts, token-by-token factorization limits parallel generation.

\tit{Continuous Diffusion for Language}
Early diffusion-based language models explore the application of diffusion processes to text by operating in continuous embedding spaces, where discrete tokens are mapped to continuous representations and corrupted with Gaussian noise before iterative denoising~\cite{li2022diffusion, gong2022diffuseq, lin2023text}. These approaches are largely motivated by the success of diffusion models in continuous domains such as image generation~\cite{ho2020denoising, song2020score}, and attempt to reuse the same iterative refinement framework within a continuous relaxation of discrete sequences.
While these approaches enable global parallel refinement, they are limited by the mismatch between continuous noise processes and discrete token semantics, as well as challenges in decoding continuous representations back into valid text.

\tit{Discrete Diffusion Language Models}
Discrete diffusion models define generative processes directly over token sequences by progressively corrupting discrete symbols and learning to reverse this corruption through iterative denoising. Foundational formulations include categorical diffusion processes and masked diffusion objectives, which generalize masked language modeling and define stochastic transition kernels over discrete vocabularies~\cite{austin2021structured, sahoo2024simple, sahoo2025diffusion}. Related approaches further connect diffusion-style denoising with energy-based and score-matching objectives in discrete settings, providing alternative views of iterative sequence reconstruction~\cite{lou2023discrete}.
A key insight in this line of work is the close relationship between autoregressive factorization and discrete diffusion processes, where autoregressive generation can be interpreted as a special case of sequential denoising under a fixed ordering. This connection has motivated methods that initialize or adapt diffusion models from pretrained autoregressive language models, enabling improved optimization and scalability in large-scale settings~\cite{gong2025scaling}.

Recent work has demonstrated that discrete diffusion can be scaled to large language modeling regimes, extending these ideas to billions of parameters and diverse text generation tasks~\cite{nie2026large, ye2025dream}. Recent studies suggest that diffusion-based language models benefit substantially from large-scale training data, exhibiting strong performance improvements as data and model size increase, as highlighted in recent large-scale analyses of diffusion language modeling~\cite{ni2025diffusion}. However, compared to autoregressive models, their scaling behavior and efficiency–quality trade-offs remain less well understood.

Despite these advances, discrete diffusion models remain computationally expensive at inference time due to the need for multiple iterations over full sequences. This iterative refinement introduces a trade-off between quality and efficiency, and makes scaling behavior less predictable compared to autoregressive models, motivating further exploration of more efficient hybrid approaches.

\tit{Block and Hybrid Diffusion Models} A recent line of work investigates block-structured diffusion models that introduce hierarchical generation schemes to improve the efficiency of diffusion-based language modeling. These approaches decompose sequence generation into coarse-grained block-level generation combined with fine-grained intra-block denoising, reducing the cost of full-sequence iterative refinement while preserving parallel token updates within blocks.

This direction is represented by a variety of block-wise diffusion formulations and hierarchical extensions that explore different strategies for partitioning sequences and scheduling denoising steps~\cite{arriola2025block, arriola2026encoder, wu2025fast}. Across these approaches, a common design principle is the introduction of structure into the diffusion process to limit the scope of iterative refinement while maintaining expressive modeling capacity.
Closely related to this family are pseudo-autoregressive diffusion models, which introduce a directional component into diffusion-based generation by iteratively refining a sliding window of future tokens conditioned on a growing prefix~\cite{liu2025sequential}. These methods blur the boundary between autoregressive decoding and diffusion by combining causal structure with iterative denoising.

Overall, these works span a spectrum of hybrid design choices that integrate autoregressive structure and diffusion-based generation at different granularities. Rather than treating generation as purely sequential or fully parallel, they explore intermediate formulations that trade off between inference efficiency, structural conditioning, and the degree of iterative refinement.

\subsection{Evaluation Protocols in Current Literature}
The development of autoregressive language modeling has been closely coupled with the establishment of standardized evaluation protocols, including unified multi-task benchmarks that enable consistent and reproducible comparison across models~\cite{biderman2024lessons, hendrycks2020measuring}. These frameworks have played a central role in ensuring that progress in autoregressive modeling is measured under comparable settings and well-defined evaluation criteria.

In contrast, diffusion-based language modeling has not yet converged on a consistent evaluation standard. Existing studies are conducted under heterogeneous experimental setups, differing in task collections, generation budgets, and sampling configurations, which limits the comparability of reported results~\cite{nie2026large, ye2025dream, wu2025fast}. As a consequence, observed performance gains are often entangled with evaluation-specific choices rather than reflecting purely architectural improvements.

This lack of standardization is particularly problematic given that diffusion models introduce an explicit computational control variable in the form of iterative denoising steps, which directly governs the trade-off between generation quality and inference cost. Without a unified protocol for varying and reporting this parameter, it remains difficult to characterize the true quality–efficiency frontier of diffusion-based language models. The issue is further exacerbated in hybrid block-diffusion settings, where additional factors such as block granularity and scheduling strategies introduce further degrees of freedom that are rarely controlled consistently across studies.

To address these limitations, this work introduces a unified experimental framework for evaluating diffusion-based language models under consistent task, budget, and inference settings, enabling systematic analysis of their quality–efficiency trade-offs.

\section{Experimental Setup}

To provide a comprehensive evaluation of DLMs, we design an experimental setup that spans both large-scale pretrained systems and controlled small-scale models trained under standardized conditions. This dual-tier structure allows us to assess performance in realistic downstream settings while also isolating architectural differences under identical data regimes.
To complement this model comparison, we evaluate performance across a diverse set of benchmarks covering knowledge, reasoning, coding, translation, and structured problem solving

\begin{table}[t]
\centering
\setlength{\tabcolsep}{.4em}
\renewcommand{\arraystretch}{1.00}
\caption{Summary of architectural and training configurations for the evaluated DLMs, categorized into large-scale and small-scale regimes. The \# Tokens column details the training data scale. Note that underlined values indicate the number of fine-tuning samples or instruction pairs, as opposed to number of tokens.}
\label{tab:dllm_specs}
\resizebox{0.97\columnwidth}{!}{%
\begin{tabular}{l ccccccc}
\toprule
\textbf{Model} & \textbf{Venue} & \textbf{\# Tokens} & \textbf{DS Training} & \textbf{Parameters} & \textbf{Block} & \textbf{Initialization} & \textbf{Denoising type} \\
\midrule
\rowcolor{gray!10}\textit{Large scale models} &&&&&&& \\
LLaDa~\cite{nie2026large}      & NeurIPS 2025 & 2.3T & Proprietary & 8B & \xmark & From Scratch & Absorbing\\
Dream~\cite{ye2025dream}      & arXiv 2025 & 580B & Open Source & 7B &  \xmark & AR & Absorbing\\
SDLM~\cite{liu2025sequential}       & arXiv 2025 & 2.3B & Open Source & 3B & \cmark & AR & Absorbing\\

LLaDa-1.5~\cite{zhu2025llada}  & arXiv 2025 & \underline{350K} & Proprietary & 8B     & \cmark & Diffusion & Absorbing\\
Fast-dLLM~\cite{wu2025fast}  & ICLR 2026 & \underline{30M} & Open Source & 7B     & \cmark & AR & Absorbing\\

\midrule
\rowcolor{gray!10}\textit{Small scale models} &&&&&&&\\
MDLM~\cite{sahoo2024simple}       & NeurIPS 2024 & 9B & OpenWebText & 200M & \xmark & From Scratch & Absorbing \\
BD3-LM~\cite{arriola2025block}     & ICLR 2025 & 9B & OpenWebText & 200M & \cmark & From Scratch & Absorbing \\
E2D2~\cite{arriola2026encoder}       & NeurIPS 2025 & 9B & OpenWebText & 170M & \cmark & From Scratch & Absorbing \\
Duo~\cite{sahoo2025diffusion}        & ICML 2025 & 9B & OpenWebText & 200M & \xmark & From Scratch & Uniform \\
\bottomrule
\end{tabular}
}
\end{table}

\subsection{Evaluated Models}

\tit{Large-Scale Downstream Models}
This tier includes recent large-scale language and diffusion models evaluated on standard reasoning, generation, and coding benchmarks, alongside autoregressive baselines.
LLaDA~\cite{nie2026large} is a large-scale discrete diffusion model trained from scratch using fully bidirectional attention and a low-confidence remasking strategy, and is evaluated under both standard full sequence diffusion and block-based sampling configurations.
LLaDA 1.5~\cite{zhu2025llada} extends this architecture with reinforcement-learning-based optimization for improved alignment and inference stability, and is evaluated in a block-diffusion setting.
Dream~\cite{ye2025dream} is a discrete diffusion language model initialized from a pretrained autoregressive checkpoint (Qwen2.5 7B). This initialization provides a strong linguistic prior, which is then adapted to iterative denoising through diffusion training.
Fast-dLLM-v2~\cite{wu2025fast} introduces a hierarchical block structure with nested sub-blocks designed to reduce inference overhead and enable more efficient sequential decoding.
SDLM~\cite{liu2025sequential} represents a hybrid formulation that combines autoregressive and diffusion-style generation by iteratively unmasking a fixed number of future tokens conditioned on a growing prefix, guided by internal confidence estimates. 
As a reference point, Qwen3~\cite{yang2025qwen3} models are included as state-of-the-art autoregressive baselines to quantify the performance and efficiency gap relative to diffusion-based approaches.

\tit{Small-Scale Controlled Models}
This tier consists of compact architectures trained from scratch on a unified corpus (OpenWebText \cite{Gokaslan2019OpenWeb}) to enable controlled comparisons under identical data conditions and allow precise perplexity evaluation without confounding factors from heterogeneous pretraining data.
MDLM~\cite{sahoo2024simple} serves as a baseline masked diffusion model that performs iterative corruption and denoising over full sequences, representing a standard discrete diffusion formulation.
BD3-LM~\cite{arriola2025block} combines autoregressive block-level generation with intra-block diffusion, enabling parallel token refinement within each segment while preserving sequential block dependencies.
Duo~\cite{sahoo2025diffusion} is a discrete diffusion model based on uniform-state corruption dynamics, leveraging a structured noise schedule that improves training stability and self-correction behavior. 
E2D2~\cite{arriola2026encoder} separates computation between an encoder processing the conditioning context and a lightweight decoder responsible for iterative denoising of target tokens.
GPT-2~\cite{radford2019language} is included as a standard autoregressive baseline trained under the same data conditions, serving as a reference point for comparing sequential and diffusion-based generation under controlled settings.

Table~\ref{tab:dllm_specs} summarizes the architectural and training characteristics of the evaluated DLMs, including parameter scale, training data, masking strategy, and whether a block-based generation scheme is employed. This overview provides a unified reference for comparing model design choices across both large-scale and controlled experimental settings.

\subsection{Datasets}

The benchmark suite includes MMLU~\cite{hendrycks2020measuring} and MMLU Redux~\cite{gema2025we} for evaluating factual knowledge and reasoning abilities, HellaSwag~\cite{zellers2019hellaswag} for commonsense reasoning and scenario completion, GSM8K~\cite{cobbe2021training} for multi-step mathematical reasoning, and HumanEval~\cite{chen2021evaluating} together with MBPP~\cite{austin2021program} for code generation. To assess performance beyond reasoning and coding tasks, we additionally include WMT16 En--De~\cite{bojar2016findings} as a machine translation benchmark and Sudoku~\cite{ye2025beyond} as a structured logical reasoning task requiring constraint satisfaction.
Together, these benchmarks provide a diverse evaluation setting covering both discriminative and generative tasks, allowing us to analyze the behavior of DLMs across a wide range of capabilities.

\subsection{Evaluation Protocol}

All experiments are conducted using the widely adopted open-source evaluation framework lm-evaluation-harness~\cite{biderman2024lessons}, which provides a standardized interface for evaluating both autoregressive and diffusion-based language models.

MMLU is evaluated in the 5-shot setting using teacher-forced log-likelihood scoring over the candidate answers, with performance reported as accuracy. HellaSwag is evaluated in the 0-shot setting using log-likelihood scoring, with performance measured through normalized accuracy to account for differences in candidate completion length and avoid bias toward shorter responses. MMLU Redux is evaluated in the 5-shot setting, where the model generates an answer and accuracy is computed from the first generated token corresponding to the predicted option label.

GSM8K is evaluated in the 4-shot setting using the flexible-extract protocol, which identifies the final numerical value in the generated response and compares it against the reference answer. HumanEval is evaluated in the 0-shot setting and MBPP in the 3-shot setting, with both benchmarks measured using pass@1 functional correctness after applying the standard output filtering procedures provided by the evaluation framework. WMT16 En--De is evaluated in the 0-shot setting using the chrF metric. Finally, Sudoku is evaluated in the 0-shot setting by verifying that generated solutions remain consistent with the input clues and satisfy all puzzle constraints.

\tit{Log-likelihood Estimation}
For all models, we compute log-likelihoods following the procedures described in their original papers. For Dream, LLaDa, and LLaDa~1.5, likelihoods are estimated through a Monte Carlo procedure, since exact autoregressive likelihoods are not directly available. For Fast-dLLM and SDLM, we follow the authors' evaluation protocol by masking all target tokens and computing the sequence log-likelihood in a single forward pass. Unless otherwise stated, we retain the original hyperparameters and scoring configurations.

\section{Large-Scale Analysis}

\subsection{Architectural Paradigm Comparison}
The results reported in Table~\ref{tab:dllm_comparison} summarize the peak performance achieved by each model under its best-performing configuration in terms of diffusion steps and block structure. This allows us to compare the intrinsic capabilities of three architectural paradigms: autoregressive modeling, full-sequence discrete diffusion, and block-based hybrid diffusion. Autoregressive models (Qwen3) are included as strong reference baselines representing standard causal language modeling, against which diffusion-based approaches are evaluated.

\tit{Pure Discrete Diffusion: Strong Knowledge Retention and Global Reasoning}
The pure diffusion models, Dream and LLaDa, exhibit markedly different profiles. Dream consistently emerges as the strongest full-sequence diffusion model, achieving the highest diffusion performance on MMLU, MMLU-Redux, HellaSwag, MBPP, and Sudoku. In particular, its $75.00\%$ Sudoku accuracy substantially exceeds both autoregressive and block-based approaches, suggesting that full-sequence iterative refinement is particularly effective for tasks requiring global constraint satisfaction. More broadly, Dream remains competitive with the larger Qwen3 baselines across several reasoning and knowledge benchmarks, indicating that diffusion-based generation can approach autoregressive performance when paired with strong initialization and sufficient inference compute.

\tit{Block-Level Diffusion: Specialization through Structured Generation}
Block-based diffusion architectures exhibit a more specialized performance profile. Fast-dLLM achieves the strongest diffusion results on GSM8K ($83.39\%$) and HumanEval ($69.51\%$), matching or surpassing much larger models on algorithmic reasoning and code generation. However, this strength comes at the expense of linguistic tasks, most notably HellaSwag, where performance drops to $30.82\%$. In contrast, LLaDa-1.5 achieves the strongest translation performance among all diffusion models ($54.85$ chrF) while maintaining competitive reasoning accuracy, suggesting that different block-generation strategies induce distinct trade-offs between sequential language modeling and structured reasoning. SDLM occupies an intermediate position, delivering competitive performance despite its smaller parameter count, particularly on coding and language understanding tasks.

Overall, the results indicate that no single diffusion paradigm dominates across all benchmarks. Full-sequence diffusion appears most effective for globally constrained and knowledge-intensive tasks, while block-based approaches can achieve superior reasoning and coding performance at the cost of greater task specialization.

\begin{table}[t]
\centering
\caption{Comparison of peak benchmark accuracy (\%) across autoregressive and DLMs. Autoregressive baselines (Qwen3) are shaded in gray, while bold text indicates the highest performance achieved among the diffusion-based architectures.}  
\label{tab:dllm_comparison}
\resizebox{\textwidth}{!}{%
\begin{tabular}{l ccccccccc}
\toprule
\textbf{Model} & \textbf{GSM8K} & \textbf{HumanEval} & \textbf{MBPP} & \textbf{MMLU} & \textbf{MMLU-Redux} & \textbf{HellaSwag} & \textbf{Sudoku} & \textbf{wmt16 en-de} \\
\midrule
\rowcolor{gray!20} Qwen3-4B   & 81.19 & 71.95 & 62.60 & 70.06 & 74.80 & 68.48 & 2.00  & 55.37 \\
SDLM       & 62.00 & 63.41 & 56.60 & 64.96 & 62.55 & 69.70 & 2.00  & 50.11 \\
\midrule
\rowcolor{gray!20} Qwen3-8B   & 87.41 & 63.41 & 64.80 & 74.78 & 78.42 & 74.97 & 8.00  & 58.21 \\

Dream      & 77.75 & 57.92 & \textbf{57.00} & \textbf{71.73} & \textbf{76.00} & \textbf{73.77} & \textbf{75.00} & 45.61 \\
LLaDa      & 65.42 & 33.53 & 40.60 & 65.84 & 61.33 & 70.91 & 46.00 & 51.31 \\
LLaDa-1.5  & 82.41 & 50.00 & 42.40 & 64.14 & 61.38 & 69.70 & 28.00 & \textbf{54.85} \\
Fast-dLLM  & \textbf{83.39} & \textbf{69.51} & 49.00 & 68.00 & 71.18 & 30.82 & 1.00  & 43.62  \\

\bottomrule
\end{tabular}%
}
\end{table}

\subsection{Scaling Analysis}

\begin{figure}[t]
  \centering
  \includegraphics[width=\linewidth]{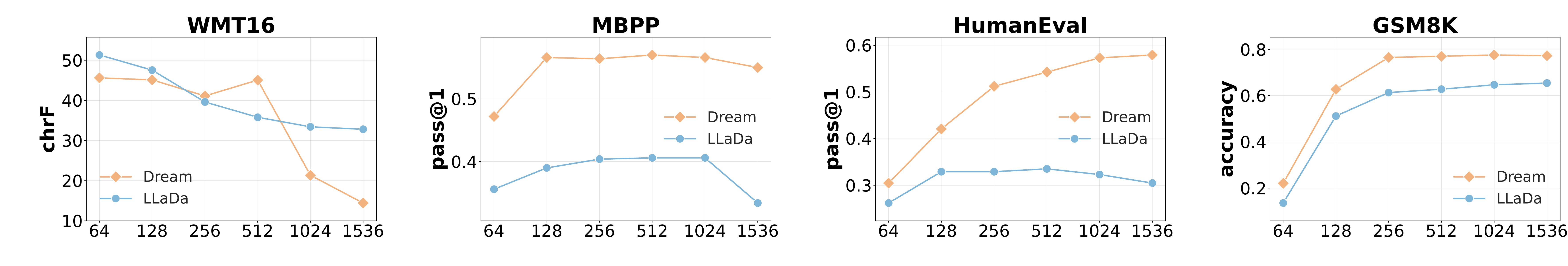}
  \vspace{-.4cm}
  \caption{Performance scaling across tasks when synchronously increasing diffusion steps and context length at a 1:1 ratio.}
  \label{fig:sync_scaling}
\end{figure}

\begin{figure}[t]
  \centering
  \includegraphics[width=\linewidth]{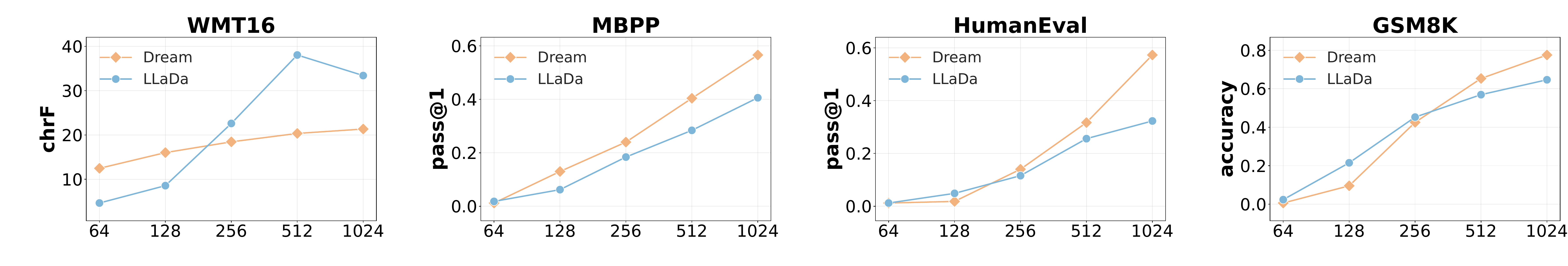}
      \vspace{-.4cm}
  \caption{Impact of scaling diffusion steps (varying the parallel unmasking ratio) while maintaining a fixed context length of $N=1024$.}
  \label{fig:parallel_unmasking}
\end{figure}

Unlike autoregressive models, DLMs expose several inference-time control variables, including the number of denoising steps, generation length, block size, and the degree of parallel token prediction. These parameters directly affect both computational cost and output quality, making them central to understanding the practical behavior of diffusion-based generation. In this section, we systematically vary these factors to characterize how different architectures utilize additional inference compute and to identify the regimes in which performance gains saturate or degrade.

\tit{Joint Scaling of Steps and Context Length}
We restrict this analysis to Dream and LLaDa, the two full-sequence diffusion models in our evaluation, in order to study scaling behavior independently of block-based generation. Figure~\ref{fig:sync_scaling} shows performance as a function of generation length, with one token denoised per step so that diffusion steps and output length scale together. For reasoning and coding tasks (GSM8K, MBPP, HumanEval), both models benefit from larger budgets initially, but performance saturates or declines beyond 256--512 tokens. The main exception is Dream on HumanEval, which continues improving at larger budgets, whereas LLaDa plateaus and eventually degrades. Translation quality (WMT16) scales poorly throughout, declining almost monotonically and collapsing at the largest budgets, suggesting that translation does not benefit from longer outputs and is particularly sensitive to error accumulation over extended generation sequences.

\tit{Scaling the Global Unmasking Ratio}
Figure~\ref{fig:parallel_unmasking} isolates the effect of the parallel unmasking ratio by fixing the generation length at N=1024 and varying the number of denoising steps. Across GSM8K, MBPP, and HumanEval, both models perform poorly at low step counts, indicating that aggressive parallel token prediction is detrimental to reasoning and code generation. Performance improves steadily as the number of denoising steps increases, with no clear saturation within the evaluated range — suggesting that in the joint scaling experiment (Figure~\ref{fig:sync_scaling}), it is the generation length rather than the step count that drives performance degradation. Dream consistently outperforms LLaDa on reasoning and coding tasks across all step budgets. Machine translation exhibits a qualitatively different pattern: both models perform poorly at low step counts, but LLaDa benefits more substantially from additional steps and eventually surpasses Dream on WMT16 at 1024 steps.

\begin{figure}[t]
  \centering
  \includegraphics[width=\linewidth]{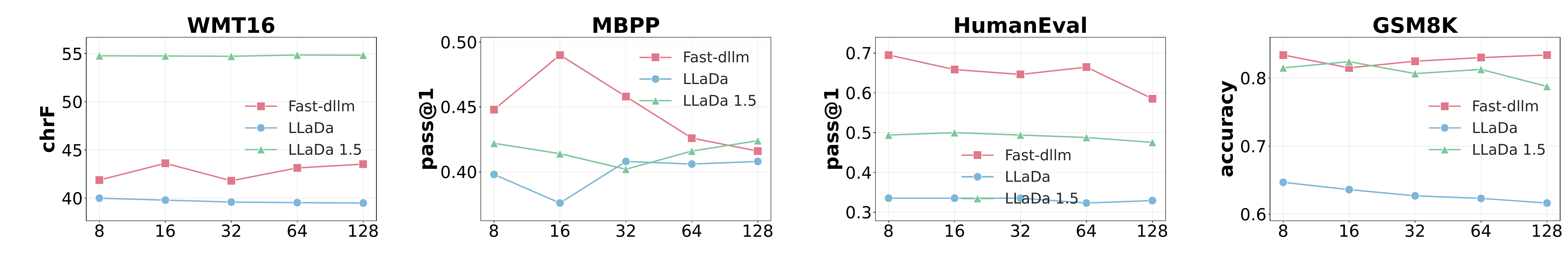}
      \vspace{-.4cm}
  \caption{Performance invariance when scaling absolute block size under a constant compute budget (1:1 ratio of generated tokens to diffusion steps).}
  \label{fig:block_invariance}
\end{figure}

\tit{Scaling Block Size under Constant Compute}
We focus this analysis on LLaDa, LLaDa 1.5, and Fast-dLLM, as these block-based models allow fixed block sizes, unlike SDLM. To isolate chunking effects, we scaled block sizes from 8 to 128 tokens while maintaining a strict 1:1 ratio of generated tokens to denoising steps. As Figure~\ref{fig:block_invariance} shows, performance is largely robust to these changes.
This stability is most evident on WMT16, where all models maintain perfectly flat curves. GSM8K and HumanEval show only minor fluctuations, implying block sizes can be flexibly tuned to meet hardware constraints (like KV-cache limits) with minimal quality degradation. MBPP is the primary exception, demonstrating higher sensitivity: Fast-dLLM spikes at block size 16, while both LLaDa models dip at smaller sizes before recovering. Regardless of these variations, relative model rankings remain strictly consistent: LLaDa 1.5 outperforms base LLaDa, and Fast-dLLM dominates reasoning and coding while trailing LLaDa 1.5 in translation.

\begin{figure}[t]
  \centering
  \includegraphics[width=\linewidth]{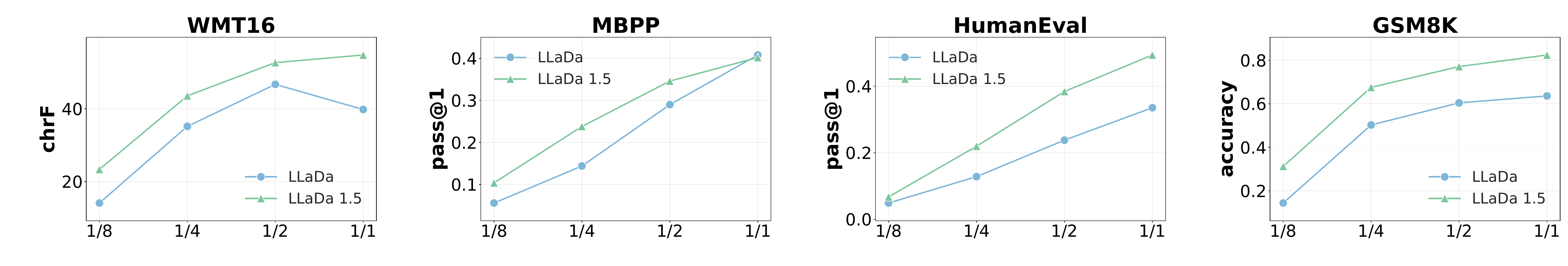}
      \vspace{-.4cm}
  \caption{Intra-block parallel unmasking effects: scaling the ratio of diffusion steps to block length within fixed block boundaries.}
  \label{fig:intra_block}
\end{figure}

\tit{Scaling the Intra-Block Unmasking Ratio}
We exclude Fast-dLLM from this analysis, as it does not allow setting a fixed amount of diffusion steps per block. To examine the cost of intra-block parallel unmasking, we scaled the ratio of denoising steps to block length (from 1/8 to 1/1) at fixed optimal block sizes. As Figure~\ref{fig:intra_block} shows, block boundaries largely preserve the qualitative behaviors seen in the global setting.
Code generation (MBPP, HumanEval) heavily penalizes intra-block parallelism: performance scales near-linearly up to 1/1, indicating fine-grained sequential refinement is necessary even within spatial chunks. LLaDa 1.5 widens its lead on HumanEval at 1/1, though both models converge identically on MBPP at this ratio. Math reasoning (GSM8K) is slightly more tolerant to parallelism; accuracy continues to grow but the rate of improvement slows noticeably after 1/4 for both models. On machine translation (WMT16), LLaDa 1.5 improves consistently across the entire range, whereas the base LLaDa model peaks at 1/2 before experiencing a distinct performance drop at 1/1.

\subsection{Computational cost analysis}

Table~\ref{tab:cost_analysis} compares the computational requirements of the evaluated DLMs in terms of peak GPU memory consumption and floating-point operations, measured both for a single forward pass and for 100 complete GSM8K generation.
For a single forward pass, Dream, LLaDa, and LLaDa~1.5 exhibit similar computational profiles, requiring approximately 16 GB of memory and 23--25 TFLOPS. In contrast, the block-diffusion architectures Fast-dLLM and SDLM are substantially more efficient, with SDLM requiring less than half the memory and compute of the pure diffusion models.

The gap widens considerably during generation. Because pure diffusion models repeatedly execute the denoising network over many refinement steps, their cumulative computational cost increases dramatically despite their parallel generation capabilities. Block-diffusion architectures mitigate this overhead by restricting diffusion to local blocks, resulting in significantly lower end-to-end generation costs.
These results highlight the fundamental quality--efficiency trade-off of DLMs: while pure diffusion architectures provide the greatest flexibility for iterative refinement, block-based approaches offer a substantially more practical inference profile.

\begin{table}[t]
\centering
\setlength{\tabcolsep}{2em}
\renewcommand{\arraystretch}{1.00}
\caption{Computational cost comparison across the evaluated models. Metrics represent peak GPU memory (VRAM) and floating-point operations (TFLOPS) for a single forward pass versus full generation on GSM8K, with autoregressive baselines shaded in gray.}
\label{tab:cost_analysis}
\resizebox{0.97\columnwidth}{!}{%
\begin{tabular}{lcccc}
\toprule
& \multicolumn{2}{c}{1 Forward Pass} & \multicolumn{2}{c}{GSM8K Generation} \\
\cmidrule(lr){2-3}
\cmidrule(lr){4-5}
Model & Peak VRAM (GB) & TFLOPS & Peak VRAM (GB) & TFLOPS \\
\midrule
\rowcolor{gray!10} Qwen3-4B & 8.36 & 5.04 & 8.36 & 9.85 \\
SDLM       & 7.49  & 4.15  & 7.51  & 6.72 \\
\midrule
\rowcolor{gray!10} Qwen3-8B & 16.68 & 9.47 & 16.68 & 19.47 \\
Dream      & 15.80 & 23.23 & 20.45 & 23726.95 \\
LLaDa      & 16.48 & 24.82 & 17.31 & 25357.94 \\
LLaDa 1.5  & 16.48 & 25.01 & 17.32 & 25557.57 \\
Fast-dLLM  & 15.50 & 9.73  & 15.55 & 33.38 \\
\bottomrule
\end{tabular}
}
\end{table}

\subsection{Small-Scale Models Analysis}

To evaluate the raw predictive confidence of compact architectures independently of specific sampling strategies, we computed perplexity on a 1000-sample ensemble dataset composed of GSM8K, MBPP, HumanEval, WMT16 En-De, MMLU, and HellaSwag.
\begin{table}[t]
\vspace{-1em}
\setlength{\tabcolsep}{2em}
\centering
\caption{Ensemble perplexity and computational cost comparison for small-scale architectures. The table reports raw predictive performance (PPL) alongside peak VRAM and TFLOPS for both a single forward pass and unconditioned generation over a fixed 1024-token sequence. The autoregressive baseline (GPT-2) is shaded in gray. }  
\label{tab:ensemble_ppl}
\resizebox{0.97\linewidth}{!}{%
\begin{tabular}{lc cccc}
\toprule
 &  & \multicolumn{2}{c}{1 Forward Pass} & \multicolumn{2}{c}{Generation} \\
\cmidrule(lr){3-4}
\cmidrule(lr){5-6}
Model  & PPL$\downarrow$ & Peak VRAM (GB) & TFLOPS & Peak VRAM (GB) & TFLOPS \\
\midrule
\rowcolor{gray!10}GPT-2  & 20.98 & 2.39 & 0.292 & 2.34 & 1.75 \\
\midrule
E2D2   & 36.82 & 5.45 & 1.03 & 5.45 & 263.15 \\
BD3-LM & 36.16 & 1.20 & 0.253 & 1.20 & 259.04 \\
MDLM   & 28.45 & 1.07 & 0.253 & 1.07 & 259.16 \\
Duo    & \textbf{24.36} & 1.08 & 0.253 & 1.08 & 259.22 \\ 
\bottomrule
\end{tabular}
}
\vspace{-1em}
\end{table}
The results reveal a clear hierarchy. GPT-2 achieves the lowest perplexity (20.98), reflecting the advantage of autoregressive models on next-token prediction. Among diffusion models, Duo obtains the strongest result (24.36), substantially improving over MDLM (28.45) and demonstrating the effectiveness of its curriculum-learning strategy. In contrast, the hybrid architectures BD3-LM (36.16) and E2D2 (36.82) exhibit considerably higher perplexities, suggesting that architectural modifications introduced to improve efficiency come at the cost of reduced likelihood modeling performance.
Computationally, MDLM, Duo, and BD3-LM exhibit nearly identical costs, whereas E2D2 requires additional memory and compute due to its encoder--decoder architecture. These differences are less pronounced than those observed in perplexity.
\section{Conclusion}

In this work, we presented a unified evaluation of modern DLMs across diverse downstream benchmarks. Our results highlight the distinct strengths and limitations of diffusion-based generation, showing that its effectiveness depends strongly on the task, inference strategy, and computational budget. While pure diffusion models benefit from global refinement, block-diffusion architectures offer a practical compromise between performance and efficiency. These findings provide a clearer understanding of the trade-offs underlying contemporary diffusion language models.

\section*{Declaration on Generative AI}
During the preparation of this work, the author(s) used ChatGPT (OpenAI) and Grammarly in order to: grammar, spelling check and rephrasing. After using these tool(s)/service(s), the author(s) reviewed and edited the content as needed and take(s) full responsibility for the publication’s content.
\clearpage
\section*{Acknowledgments}
This work has been supported by EU Horizon project ELLIOT (No. 101214398) and by the EuroHPC JU project IT4LIA (No. 101234224). We also acknowledge CINECA for the availability of high-performance computing resources under the ISCRA initiative.

\bibliography{sample-ceur}

\end{document}